\documentclass[conference]{IEEEtran}
\IEEEoverridecommandlockouts
\usepackage{cite}
\usepackage{amsmath,amssymb,amsfonts}
\usepackage{algorithmic}
\usepackage{graphicx}
\usepackage{textcomp}
\usepackage{xcolor}

\usepackage{booktabs}
\usepackage{graphicx}

\newcommand{\bm}[1]{\boldsymbol{#1}}

\def\BibTeX{{\rm B\kern-.05em{\sc i\kern-.025em b}\kern-.08em
    T\kern-.1667em\lower.7ex\hbox{E}\kern-.125emX}}
\begin{document}



\makeatletter
\newcommand{\linebreakand}{%
  \end{@IEEEauthorhalign}
  \hfill\mbox{}\par
  \mbox{}\hfill\begin{@IEEEauthorhalign}
}
\makeatother

\title{Uncovering the Folding Landscape of RNA Secondary Structure Using Deep Graph Embeddings \\
\thanks{$^\dagger$ Equal contribution, corresponding authors. This research was partially funded by: IVADO Professor startup \& operational funds, IVADO Fundamental Research Project (grant PRF-2019-3583139727) [\emph{G.W.}]; Chan-Zuckerberg Initiative grants 182702 \& CZF2019-002440 [\emph{S.K.}]; and NIH grants R01GM135929 \& R01GM130847 [\emph{G.W., S.K.}]. The content provided here is solely the responsibility of the authors and does not necessarily represent the official views of the funding agencies.}
}

\author{\IEEEauthorblockN{Egbert Castro}
\IEEEauthorblockA{\textit{Comp. Biol. and Bioinf. Program} \\
\textit{Yale University}\\
New Haven, CT, USA \\
egbert.castro@yale.edu}
\and
\IEEEauthorblockN{Andrew Benz}
\IEEEauthorblockA{\textit{Dept. of Mathematics} \\
\textit{Yale University}\\
New Haven, CT, USA \\
andrew.benz@yale.edu}
\and
\IEEEauthorblockN{Alexander Tong}
\IEEEauthorblockA{\textit{Dept. of Computer Science} \\
\textit{Yale University}\\
New Haven, CT, USA \\
alexander.tong@yale.edu}
\linebreakand
\IEEEauthorblockN{Guy Wolf$^{\dagger}$}
\IEEEauthorblockA{\textit{Dept. of Math. and Stat.} \\
\textit{Univ. de Montr\'{e}al; Mila} \\
Montreal, QC, Canada \\
guy.wolf@umontreal.ca}
\and
\IEEEauthorblockN{Smita Krishnaswamy$^{\dagger}$}
\IEEEauthorblockA{\textit{Depts. of Genetics \& Comp. Sci.} \\
\textit{Yale University} \\
New Haven, CT, USA \\
smita.krishnaswamy@yale.edu}
}



\IEEEoverridecommandlockouts
\IEEEpubid{\makebox[\columnwidth]{978-1-7281-6251-5/20/\$31.00~\copyright2020 IEEE \hfill} \hspace{\columnsep}\makebox[\columnwidth]{ }}

\maketitle

\begin{abstract}

Biomolecular graph analysis has recently gained much attention in the emerging field of geometric deep learning. Here we focus on organizing biomolecular graphs in ways that expose meaningful relations and variations between them. We propose a geometric scattering autoencoder (GSAE) network for learning such graph embeddings. Our embedding network first extracts rich graph features using the recently proposed geometric scattering transform. Then, it leverages a semi-supervised variational autoencoder to extract a low-dimensional embedding that retains the information in these features that enable prediction of molecular properties as well as characterize graphs. We show that GSAE organizes RNA graphs both by structure and energy, accurately reflecting bistable RNA structures. Also, the model is generative and can sample new folding trajectories.
\end{abstract}


\section{Introduction}

An emerging focus in deep learning is the ability to analyze graph structured data. Historically, these types of data have mainly originated from network analysis fields, such as the study of social networks or citation networks. \cite{perozzi2014deepwalk, hamilton2017inductive, jiang2019semi}. More recently, interest in graph data analysis has also risen in the equally important study of \textit{biomolecules}, represented by graphs that model their molecular structure. For this work, we define \textit{biomolecules} as chemical and biological entities that form structure though atomic interactions, such as small-molecules, proteins, and RNA. This data domain presents both great challenge and opportunity. The challenge arises from the (1) the discrete nature of graph structure data, (2) the vast space of possible structures possible ($10^{30} -10^{60}$ drug-like molecules for example \cite{polishchuk2013estimation}), and (3) the underlying physical constraints for validity. However, advancements in this domain have the potential for greater insights into biological questions as well as improvements in drug discovery.

Here we focus on the data domain of RNA secondary structures. While RNA is sometimes thought of as a linear sequence of bases, non-coding RNA especially can fold into 3D structures that possess important functionality within the cell \cite{ganser2019roles}. Each RNA sequence has the propensity of folding into many different structures transiently, but fewer structures stably. Though methods exist to predict the minimum free energy (MFE) structure from sequence \cite{zuker1981optimal,nussinov1978algorithms}, the reliance of these methods on simple energy approximations often results in the proposed structure being a poor prediction of the \textit{in-vivo} dominant fold. Rather, it helps to explore the functionality of RNA structures if we can embed them in ways that uncover features of their folding landscape and smoothly reflect their transitions. This motivates the examination of graph embeddings generated through combining graph scattering transforms and neural networks to see if they can organize RNA graphs into coherent landscapes or folding manifolds. Using such embeddings, biologists could, for example, assess the likelihood that the RNA could switch from one structure to another to change functionality, i.e., whether it is a riboswitch -- a question that is difficult to answer normally \cite{schroeder2018challenges}. 

Our embedding procedure based on three goals: $\bullet$ Obtaining faithful embeddings where neighbors are close in both in terms of graph structure and in terms of molecular properties.
$\bullet$ Enabling visual exploration and interpretation of biomolecular structures using the embedding space.
$\bullet$ Generating trajectories of folds and decode them into molecular folds. We define these desiderata more formally in Section \ref{sec:setup}.

Here, we propose a new framework for organizing biomolecular structures called geometric scattering autoencoder (GSAE). First, GSAE encodes molecular graphs based on scattering coefficients \cite{pmlr-v97-gao19e, gama2019stability} of dirac signals placed on their nodes. Next, it uses an autoencoder architecture to further refine and organize the scattering coefficients into a reduced and meaningful embedding based on both a reconstruction penalty and auxiliary penalties to predict molecular properties. Finally, to generate new graphs we train a scattering inversion network (SIN) that takes scattering coefficients as inputs and generates adjacency matrices. 

The main contributions of this work are as follows:
 
\begin{itemize}
\item  We propose the Graph Scattering Autoencoder (GSAE) that combines graph scattering transforms with a variational autoencoder framework to learn rich representations of biomolecules.
\item We demonstrate that GSAE faithfully embeds ensembles of RNA molecular folds as well as synthetic graphs. 
\item We propose a scattering inverse network, which to our knowledge is the first systematic inversion of the scattering transform and allows for the generation of quasi-trajectories in the RNA folding domain.

\item We compare our results to several of the most prominent GNN-based graph representation approaches including GAE \cite{kipf2016variational}, GVAE \cite{kipf2016variational}, as well as non-trainable methods like embeddings of the WL-kernel computed on graphs \cite{shervashidze2011weisfeiler} or embeddings of graph edit distance matrices on toy and RNA datasets.
\end{itemize}

\begin{figure}[t]
    \centering
    \includegraphics[width=0.95\linewidth]{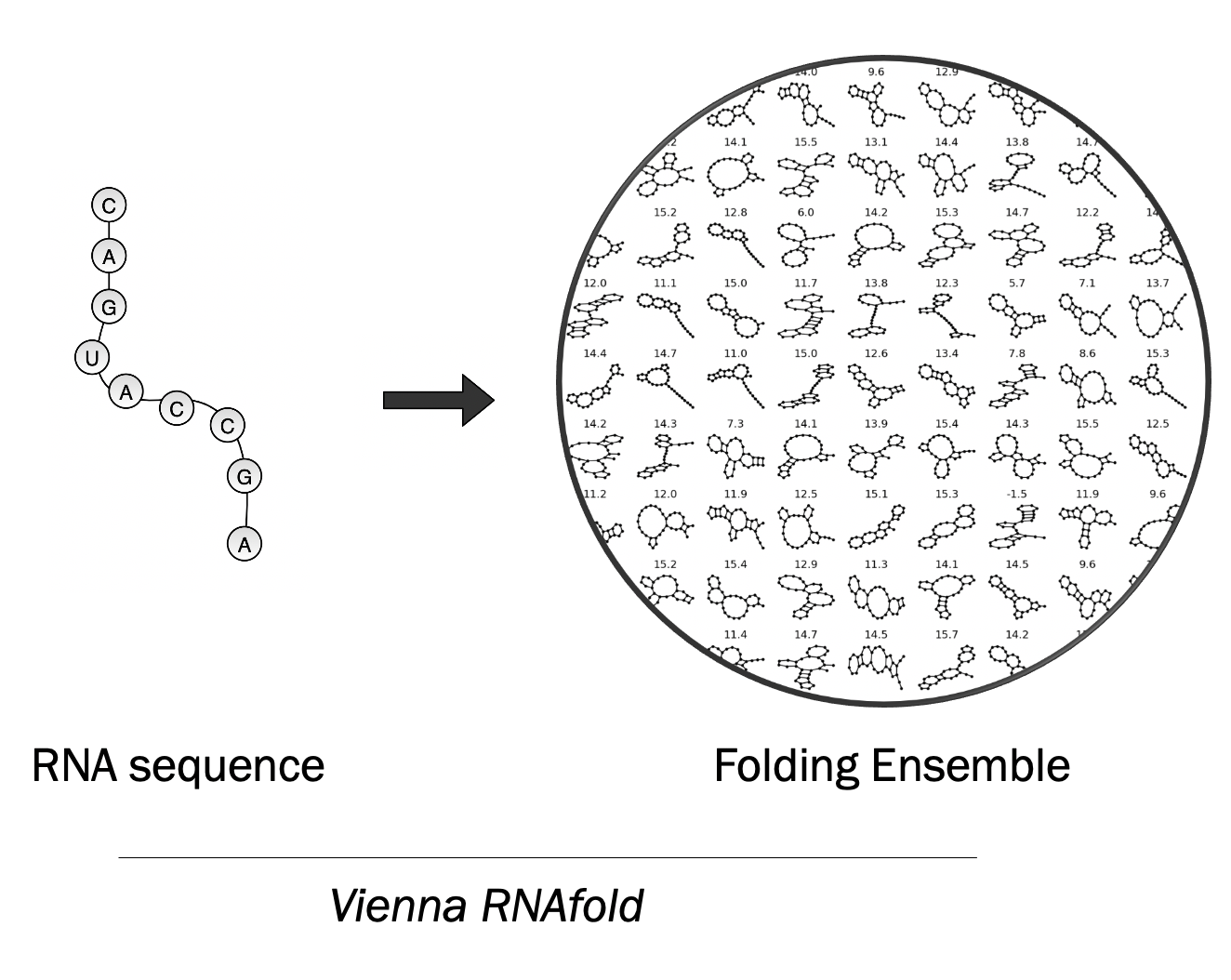}
    \caption[textfont=small]{To look beyond the predicted minimum free energy (MFE) structure of a sequence, we instead generate an ensemble of "suboptimal" structures that provide a broader view of the structural diversity possible. For this study, we rely on a software package called Vienna RNAFold}
    \label{fig:seq2ensemble}
\end{figure}

\section{Background}

\subsection{RNA Structure}

Difficulty in experimentally determining the structure of RNA molecules \cite{zhang2019new} has drawn efforts towards computational approaches to RNA secondary structure prediction. Dynamic programming algorithms, such as Zuker's \cite{zuker1981optimal} or Nussinov's algorithm \cite{nussinov1978algorithms}, search for the minimum free energy (MFE) structure and are among the most popular approaches to predicting RNA secondary structure from sequence. However, this focus on the MFE draws attention away from structures whose energies may be slightly above that of the MFE and may hinder novel findings in this domain for two reasons. The first of which is that the dynamic and crowded intracellular environment is likely to prevent RNA from folding into its MFE \cite{zou2008review}. Secondly, inaccuracies in the energy function used by the RNA secondary structure prediction method may produce a false MFE. Notably, the Zuker algorithm's accuracy drops as sequences grow in length. For these reasons it is vital for one to examine the larger set of folds possible to fully understand the structural diversity of a given sequence, depicted in Figure \ref{fig:seq2ensemble}.

Like all biomolecules, RNA exists in its lowest order structure as a chain of monomers. This sequence of units, called nucleotides, which are able to interact with each other to form intra-chain hydrogen bonds. The collection of these interactions can be readily interpreted as edges between the nucleotides, which consequently form the node set. In this way, we can view the secondary structure of RNA as a graphs.  This viewpoint allows for the application of existing tools from both graph signal processing and graph neural networks to be leveraged in this domain. 

\subsection{Geometric Scattering}

The geometric scattering transform~\cite{pmlr-v97-gao19e,gama2019diffusion} is based on a cascade of graph wavelets, typically constructed via diffusion wavelets~\cite{COIFMAN200653}. These are constructed using a lazy random walk diffusion operator $\mathbf{P} = \frac{1}{2} ( \mathbf{I} + \mathbf{A}\mathbf{D}^{-1})$, where $A$ is the adjacency matrix of the analyzed graph and $D$ is a diagonal matrix of its vertex degrees. Then, $P^t$, $t > 0$, contains $t$-step diffusion transition probabilities between graph nodes. These powers of $P$ can also be interpreted as lowpass filters that average signals over multiscale diffusion neighborhoods, where the size (or scale) of the neighborhood is determined by $t$. Therefore, given a graph signal $f$, the filtered signal $\mathbf{P}^t f$ only retains intrinsic low frequencies over the graph. Similarly, $I - P^t$, $t >0$, form a highpass filters whose scales is determined by $t$. The diffusion wavelets transform~\cite{COIFMAN200653} combines these lowpass and highpass filters to form bandpass filters of the form $\bm{\Psi}_j = \mathbf{P}^{2^{j-1}} - \mathbf{P}^{2^j} = \mathbf{P}^{2^{j-1}} (\bm{I} - \mathbf{P}^{2^{j-1}})$, with dyadic scales $2^j$, $j = 1, \ldots, J$ where $J$ defines the widest scales considered (corresponding to $2^J$ random walk steps). The resulting wavelet transform then yields wavelet coefficients $\bm{\mathcal{W}} f = \{ \mathbf{P}^t f, ~ \bm{\Psi}_j f \}_{j=1}^{\log_2 t}$ that decompose $f$ into a family of signals that capture complementary aspects of $f$ at different scales (i.e., intrinsic frequency bands on the graph).

While the wavelet coefficients $\bm{\mathcal{W}}f$ give a complete and invertible representation of $f$, the representation provided by $\bm{\Psi}_j f$ is not guaranteed to provide stability or invariance to local deformations of the graph structure. To obtain such representation, \cite{pmlr-v97-gao19e} propose to follow the same approach as in expected scattering of traditional signals~\cite{mallat2012group,bruna2013invariant} to aggregate wavelet coefficients by taking statistical moments after applying nonlinearity in form of absolute value. Their first-order scattering features are $ S_1 f = \left [ \|\, |\bm{\Psi}_j f| \, \|_q \right]_{1 \leq j \leq J, 1 \leq q \leq Q}$, which capture the statistics of signal variations over the graph. They are complimented on one hand by zeroth-order scattering, consisting of statistical moments of $f$ itself (without filtering), and on the other hand with higher order scattering coefficients that capture richer variations eliminated by the aggregation in the above equation. In general, $m^{\text{th}}$ order scattering features are computed by a cascade of $m$ wavelet transforms and absolute-value nonlinearities, creating a designed (i.e., non-learned) multiscale graph neural network:  $S_m [j_1, \ldots, j_m,q] f = \| \, |\bm{\Psi}_{j_m} | \cdots | \bm{\Psi}_{j_1} f| \cdots || \, \|$,  with features indexed by moment $q$ and scales $j_1, \ldots, j_m$. Due to the multiresolution nature of these features, they provide a rich and stable description of $f$~\cite{pmlr-v97-gao19e, perlmutter2019understanding, gama2019stability, gama2019diffusion}

\section{Related Work on Graph Embeddings}

Graph edit distances (GED) are a way of measuring the distances between graphs based on the number of elementary operations needed to change from one graph to another. These elementary operations involve vertex insertions and deletions, edge insertions and deletions, etc. Distances can directly be embedded using MDS or indirectly via a Gaussian kernel using a kernel-PCA. Methods such as diffusion maps~\cite{coifman2006diffusion} or the more recently proposed PHATE~\cite{moon2019visualizing} collect manifold information for visualization in two dimensions. Another approach to embedding a graph is the Weisfeiler-Lehman (WL) kernel~\cite{shervashidze2011weisfeiler} which maps a graph to a sequence of graphs that encapsulate graph topological features. 

Though graph neural networks have been primarily used for node classification, methods such as graph autoencoders (GAEs)~\cite{kipf2016variational} have been used for embedding nodes and graphs. To achieve a graph represenation that is invariant to node orderings, node embeddings must undergo a pooling step. Similar to convolutional neural networks in the image domain, graph neural networks typically use sum or max pooling approach \cite{hamilton2017representation}. Here, inspired by deep scattering transforms~\cite{pmlr-v97-gao19e}, we instead use the statistical moments of node activations for pooling. 

There has been much work in recent years using graph-based methods for a related class of biomolecules, commonly referred to in the literature as \textit{small molecules}. For a review of these methods we refer the reader to \cite{elton2019deep}. In this related domain, a similar approach to organizing biomolecules in latent space using an auxiliary loss has been previously studied by \cite{gomez2018automatic}. However their approach relies on RNNs to encode and decode a domain-specific string representation for this class of biomolecules rather than molecular graphs directly.

In regards to RNA secondary structures, few experiment-free, graph-based approaches to interpreting a RNA secondary structure folding ensemble (a collection of folds arising from a single sequence) have been studied. The popular RNAShapes software seeks to abstract the structural diversity of a folding ensemble in a coarser set of possible graphs \cite{steffen2006rnashapes}.  MIBPS is a method which utilizes mutual information between folds of an ensemble to predict multi-modality \cite{lin2018characterization}. Furthermore, several works rely on chemical probing data in order to infer multi-modality in folding ensembles \cite{cordero2015rich} \cite{siegfried2014rna},\cite{spasic2018modeling}, \cite{woods2017comparative}. Closer to the deep learning literature is a recent work by  \cite{yan2020graph} where the authors train a GNN-based model to study RNA secondary structures in the context of RNA binding proteins (RBPs).

\section{Methods}
\label{sec:methods}

\subsection{Problem Setup} \label{sec:setup}
Given a set of graphs $\mathcal{G}=\{G_1, G_2, \ldots, G_n\}$, we aim to find an embedding $Z_{\mathcal{G}}=\{z_1,z_2,\ldots, z_n\}$ in Euclidean space, i.e., where each graph $G_i$ is mapped to a $d$-dimensional vector $z_i \in \mathbb{R}^d$, where the embedding satisfies the following properties, which we will validate empirically for our proposed construction:
\begin{enumerate}
\item \textbf{Faithfulness:} the embedding should be faithful to the graphs in $G$ in the sense that graphs that are near each other in terms of graph edit distance should be close to each other in the embedding space, and vice versa. Formally we aim for $\|z_i-z_j\|<\epsilon$, for some small $\epsilon$, to be (empirically) equivalent to $ged(G_i, G_j)<\nu$ for some small $\nu$ where $ged$ is graph edit distance.  
\item \textbf{Smoothness:} the embedding should be smooth in terms of a real valued meta-property $M=\{m_1, m_2, \ldots, m_n\}$, where $m_i \in \mathbb{R}^n$, which is only given on the training data.
\item \textbf{Invertibility:} it should be possible to generate new graphs by interpolating points in the embedded space and then inverting them to obtain interpolated graphs between training ones. Formally, for any two points $z_x,z_y$ in the embedding space, we expect $z = (z_x+z_y)/2$ to match the embedding of a valid graph, with properties specified in the previous criteria, and with a constructive way to (approximately) reconstruct this graph. 
\end{enumerate}

To explain the second criterion, given an affinity matrix of vectors in $Z_{\mathcal{G}}$, denoted $A_{Z_{\mathcal{G}}}$, where $A_{\mathcal{G}} (i,j)= similarity(z_i, z_j)$, we define a Laplacian matrix of this embedding as $L=D-A_{\mathcal{G}} $ where $D$ is a diagonal matrix whose entry $D(i,i)=\sum_j A(i,j)$, we want the {\em dirichlet energy} $M^T L M$ to be small. However, the difficulty in biological graphs is that $M$ is an emergent property that can be difficult to compute from the graph. In principle, this smoothness could be enforced for multiple meta properties. 

\subsection{Geometric Scattering Autoencoder}\label{sec:geom-scat}

To derive an embedding that has the properties described in the previous section, we propose a novel framework based on the untrained geometric scattering, a trained autoencoder, and a scattering inversion network, as shown in Figure~\ref{fig:schematic}. 

\begin{figure}[!ht]
    \centering
    \includegraphics[width=0.95\linewidth, height=6cm]{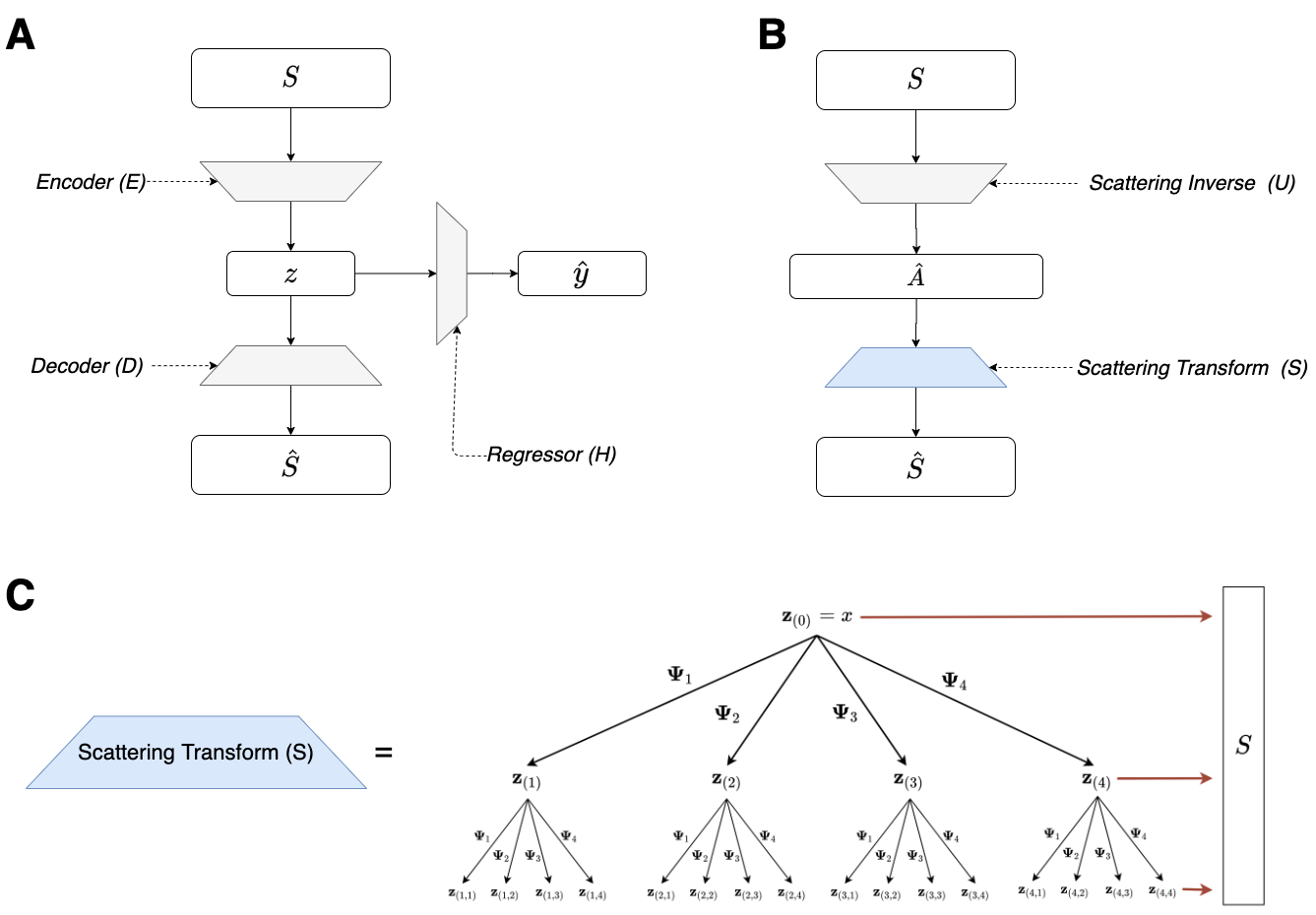}
    \caption[textfont=small]{A. GSAE, B. Inverse transform transform network. C. Scattering Transform network (S)}
    \label{fig:schematic}
\end{figure}

The first step in our construction is to extract scattering features from an input graph, thus allowing us to further process the data in a Euclidean feature space. Since the biomolecule graphs considered in this work do not naturally provide us with graph signals, we have to define characteristic signals that will reveal the intrinsic graph structure. However, since we mostly focus here on RNA folding applications, we assume there is node correspondence between graphs, and thus we can produce a set of diracs $d_i = \{0, \ldots, 1, \ldots, 0\}$ that provide one-hot encoding of each node $v_i$ in the graph (i.e., $d_i[j] = 1$ iff $i=j$; zero otherwise). 

Next, we map an input graph to a Euclidean feature space given by the scattering features of these dirac signals over the graph. For each dirac signal $d_i$, we take the zeroth, first and second order scattering and concatenate them across orders. Then, we concatenate the scattering coefficients of all the dirac signals over the graph to obtain its entire scattering feature vector. Formally, let $\Pi$ denote the concatenation operator, then this feature vector is given by $S(G) = \Pi_{i=1}^n \Pi_{m=0}^2 S^{(G)}_m d_i$ is constructed using graph wavelets from a lazy random walk over the graph $G$, where the superscript indicates the scattering operation.

The scattering representation provided by $S(G)$ encodes the graph geometry in a Euclidean feature space that is high dimensional and often highly redundant. Indeed, as shown in \cite{pmlr-v97-gao19e}, it is often possible to reduce significantly the dimensionality of scattering representations while still maintaining the relations between graphs encoded by them. Therefore, the next step in our embedding construction is to apply an autoencoder to the scattering features in $S(G)$. Formally, we train an encoder $E(\cdot)$ and decoder $D(\cdot)$ such that $\hat{S}(G) = D(E(G))$ will approximately reconstruct $S(G)$ via a MSE penalty $\|S(G) - \hat{S}(G)\|^2$. However, as mentioned in Sec.~\ref{sec:setup}, in addition to the unsupervised information captured and provided by $S(G)$, we also aim for our embedding to follow physical properties of the biomolecules represented by the graphs. These are encoded by meta properties available at the graph level, denoted here by $m(G)$. Therefore, in addition to the reconstruction penalty, we also introduce a supervised penalty in the loss for predicting $m(G)$ via an auxiliary network $H(\cdot)$ operating on the latent embedding. Formally, this penalty is added to the autoencoder loss via a term $\|m(G) - H(E(S(G)))\|^2$

Finally, since we aim for our embedding to be approximately invertible, we must also construct a transform that maps embedded representations into viable graphs. We recall that our data consists of graphs that all share the same nodes, and therefore this construction is only required to infer an adjacency matrix from embedded coordinates. The autoencoder trained in the previous step naturally provides a decoder that (approximately) inverts the latent representation into geometric scattering features. Furthermore, to ensure stability of this inversion to perturbation of embedded coordinates, as well as enable (re)sampling from the embedding for generative purposes, we add a VAE loss term to our autoencoder, injecting noise to its latent layer and regularizing its data distribution to resemble normal distribution via KL divergence as in \cite{kingma2013auto}. 

Our final step is to construct a scattering inversion network (SIN) that is able to construct adjacency matrices from scattering features. We observe that the main challenge in optimizing such an inversion network is how to define a suitable loss on the reconstructed adjacency matrices. We mitigate this by leveraging the geometric scattering transform itself to compute the inversion loss. Namely, we treat the concatenated construction of $S(\cdot)$ as a decoder and then train the inversion network $U(\cdot)$ as an encoder applied to $(G)$ such that the scattering features of the resulting graph will approximate the input ones, penalized via the MSE: $\|S(G) - S(U(S(G))\|^2$.

Putting all the components together, the geometric scattering autoencoder (GSAE) trains four networks ($E$,$D$,$H$,$U$) with a combined loss: $\mathbb{E}_{G \in \mathcal{G}} \|D(E(S(G))) - S(G)\|^2 + \alpha \|H(S(G)) - m(G)\|^2 + \beta \|S(G) - S(U(S(G))\|^2$, where $\alpha$ and $\beta$ are tuning hyperparameters controlling the importance of each component in the loss.

\subsection{Smoothness Metric}

As we are interested in enforcing the energy smoothness property described in Section \ref{sec:setup}, we first describe a metric from which we quantify the smoothness of a signal in embedding space. This is done using graph dirichlet energy. This metric can be interpreted as the squared differences between neighboring nodes, which should be small if the signal is smooth and slow varying across latent space. Conversely, large differences in the quantity of interest between neighboring nodes would produce as large value of this metric. Here we use a normalized form of the graph dirchlet energy, described in \cite{dakovic2019local} as a smoothness index, which takes the form, 

$$ \lambda_{m}=\frac{\mathbf{m}^{T} \mathbf{L} \mathbf{m}}{\mathbf{m}^{T} \mathbf{m}}$$

Computing graph dirchlet energy requires that we first form a graph on our embeddings in order to compute the graph Laplacian $\textbf{L}$. We do this using a symmetric k-nearest neighbor (kNN) graph where a data points $x_i$ and $x_j$ are connected by an unweighted edge in the graph if either $x_i$ or $x_j$ fall within each other's kNN. Nearest neighbers are determined using Euclidean distance between points in latent space. In our work we set k equal to 5 unless otherwise stated.

\subsection{Extracting Graph Embeddings frm GSAE}

In this work we begin with graphs $G$, that generally have no features. To derive an embedding of such graphs, which use a set of $n$ artificial features, which consist of diracs placed on each node. These will allow the GSAE to capture the structure of the graph by computing the way these signals are scattered.   We then generate a set of node features using the scattering transform formulation depicted in Figure \ref{fig:schematic} and described in Section \ref{sec:geom-scat}. To achieve a graph representation, we summarize node features using statistical moments from \cite{pmlr-v97-gao19e} rather than the traditional sum or max operation used in most graph autoencoders. We refer to this graph representation as $S$.

The autoencoder takes as input the summarized scattering coefficients, $S$. In the GSAE model, shown in Figure \ref{fig:schematic}, we use 2 fully-connected layers with RELU activations followed by the reparameterization operation described in \cite{kipf2016variational}. Batchnorm layers from \cite{ioffe2015batch} are interspersed between the initial encoding layers.  The decoder of GSAE is comprised of 2 fully-connected layers with a RELU activation function on the non-output layer. For the regressor network, we an identical module as the decoder, only differing the size of the output layer. The loss which is optimized during training becomes,

$$L = L_{recon} +  \alpha L_{pred} + \beta L_{D_{KL}} $$

or,

$$L = \frac{1}{N} || \hat{\Phi}, \Phi ||_2^2+  \alpha \, \frac{1}{N} ||\hat{m} - m ||_2^2 +  \beta \, D_{\mathrm{KL}}\left(q( z | \Phi ) \| p(z )\right) $$

Training runs consisted of 15000 iterations using a batch size of 100. We used PyTorch's Adam optimizer with a learning rate of 0.0001. For experimental results presented we use a bottleneck dimension of 25, unless otherwise stated.

\subsection{Generative modeling using the Scattering Inverse Network Model}
\label{scat-inverse}

From the GSAE, we are able to produce a latent space where both information about graph structure and graph metaproperties are preserved. However the GSAE construction differs from other graph autoencoders as it reconstructs summarized scattering coefficients rather than graphs. This presents an obstacle when generating graphs from points in the latent space. We remedy this by training an additional model referred to as the Scattering Inverse Network (SIN) model. 

Similar to GSAE, SIN uses an autoencoder architecture which reconstructs scattering coefficients. However SIN differs from GSAE as it produces the graph adjacency matrix in it's middle latent representation. This endows SIN with the capacity to effectively invert scattering coefficients and consequently, allow for generation of graphs from the GSAE's latent space. 

For SIN, depicted in Figure \ref{fig:schematic}B, we use 2 blocks of fully-connected layer $\Rightarrow$ RELU $\Rightarrow$ Batchnorm followed by a final fully-connected layer. This final fully-connected layer expands the representation  so that the inner-product decoder of GAE \cite{kipf2016variational} may be applied to produce an adjacency matrix representation of the graphs. Unique to SIN is that we then convert the adjacency matrix to scattering coefficients $\hat{S}$ using the original scattering cascade  used to construct the input to GSAE. 

We train SIN by first pre-training the scattering inverse module which takes $S$ to $\hat{A}$ using a binary-cross entropy loss. Once this loss has converged, we then refine the generator by training on the overall reconstruction of $S$. We show these final MSE losses for the RNA datasets in Table \ref{table:inv-quant}.

\section{Empirical validation}
In this section we validate the GSAE and SIN models empirically using toy and RNA datasets, via rigorous comparisons to other models on the criteria of faithfullness, metaproperty smoothness and invertibility. set up in Section \ref{sec:setup}.  We compare GSAE against traditional graph autoencoders, the GAE and GVAE from \cite{kipf2016variational}. Though more complex graph autoencoders have been developed for domain-specific applications (e.g. small molecules from chemistry), we focus on a more general sense of graph embeddings which do not rely on existing node features but rather only utilize graph structure and an associated meta-property. 

To make set-up as similar to GSAE as possible, we again begin with featureless graphs $G$ on which we place diracs as the initial node signal. The GAE and GVAE both use this initial signal to create meaningful node features using graph convolutional (GCN) layers  from \cite{kipf2016semi}. In this work we use 2 GCN layers with RELU activation functions for both GAE and GVAE. We then attain a graph-wise representation using the same pooling as GSAE, which uses the first 4 statistical moments across the node dimension. The resulting vector is then passed through two fully-connected layers to produce the final latent representation which is used for evaluations. We train these models using a binary-cross entropy loss for 15000 iterations with batch size set to 100. As with GSAE, we use PyTorch's Adam optimizer with a learning rate of  0.0001.

\paragraph{Ablation Study} We also train three variations of the GSAE itself to achieve an ablation study. First we examine the effect of the variational formulation by including results for GSAE-AE (our model trained as a vanilla autoencoder). We also truncate the regressor network \textit{H} of GSAE, which we refer to as GSAE (without the H). We compare and show GSAE also improves upon simply embedding geometric scattering coefficients, which may contain information for organizing the graphs, but are not selected, weighted or combined as well as in the proposed GSAE. Finally, we identify that the statistical pooling mechanism described in Section \ref{sec:geom-scat} is an important part of the increased richness of the scattering coefficients, and we show the effect of incorporating this pooling rather than max or sum pooling in the other graph autoencoders.

\subsection{Toy Dataset Description}

\begin{figure}[!ht]
    \centering
    \begin{tabular}{cc}
    \includegraphics[width=6cm,height=6cm]{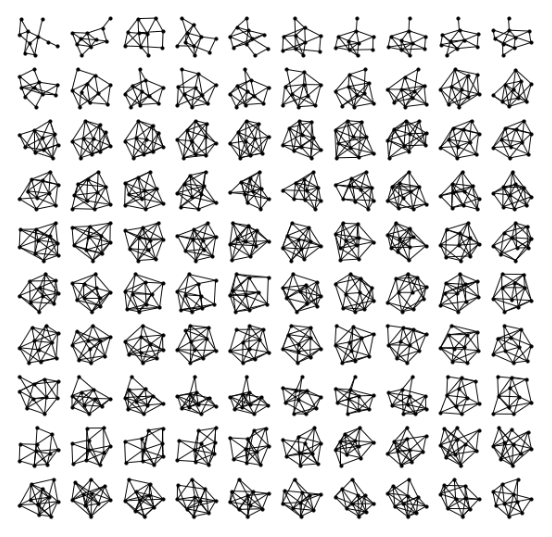} 
    \end{tabular}
    \caption[textfont=small]{First 100 steps in the toy graph trajectory}
\end{figure}

For evaluation of our model on a noise-less toy dataset, we create a graph trajectory starting from an initial Erdős-Rényi or binomial graph with p=0.5. Each step in this trajectory is either an edge addition or deletion. Starting from the initial graph, we take 9999 steps and save each step's graph. After the final step, we have produced a sequence of graphs which we refer to as a trajectory. Here we define the metaproperty of interest, $m$ as the step index of of a graph with respect to the trajectory. That is, $m_i$ of graph $G_i$ corresponds to $G_i$'s position in the overall 10000 step sequence of graph edits.

\subsection{RNA Dataset Description}

The four RNA secondary structure datasets used in this work were generated using ViennaRNA's RNAsubopt program \cite{lorenz2011viennarna}. For these datasets, we are interested in the metaproperty that corresponds to a structures stability, namely it's Gibbs free energy in units of kcal/mol. The Vienna RNAfold program takes as input an RNA sequence and produces a set of folds and their associated stabilities. This program performs dynamic programming to exhaustively sample structures within an energy range and returns an approximate energy for each structure. Here we used the "-e" option which produces an exhaustive set of folds within a specified kcal/mol energy range above the minimum free energy (MFE) structure. We then split each dataset into a train and test split with a ratio of 70:30.  For the purpose of testing embedding quality we chose four sequences that were identified as having specific structures in literature,

\begin{itemize}
    \item \textbf{SEQ3}: SEQ3 is an artificial RNA sequence of 32 nucleotides designed to be bistable \cite{hobartner2003bistable}.  We use an energy window of 25kcal/mol which produces a total of 472859 sequences. We then reduce this set to 100k structures by sampling without replacement.
    \item \textbf{SEQ4}: SEQ4 is also an artificial RNA sequence of 32 nucleotides and is bistable \cite{hobartner2003bistable}. We use a 30kcal/mol window which produces 926756 structures. We then reduce this set to 100k structures by sampling without replacement.
    \item \textbf{HIVTAR}: HIVTAR \cite{ganser2019roles} refers to the ensemble generated from the transactivation response element (TAR) RNA of HIV. It has been used as a model system for studying RNA structural dynamics and is one of the few RNAs with single native secondary that dominates. HIVTAR is 61 nucleotides long and we use a 22kcal/mol window which produces 1529527 structures.  We then reduce this set to 100k structures by sampling without replacement.
    \item \textbf{TEBOWN}: TEBOWN was designed to be bistable but was described as a "faulty riboswitch" \cite{cordero2015rich}, displaying 3 or more dominant states. TEBOWN has a sequence length of 72 nucleotides and is expected to be multistable. We use a 9kcal/mol window which produces 151176 structures. We then reduce this set to 100k structures by sampling without replacement.
\end{itemize}

\subsection{Toy Dataset Experiments}

Using the toy dataset described above, we examine the ability of the non-trained and trainable models to recover a trajectory in their embedding. These toy graphs are visualized in Figure~\ref{fig:toy}A. We visualize these embeddings in two different ways, with PHATE  \cite{moon2019visualizing} a non-linear visualization reduction method that keeps local and global structure,  as well as PCA. We see that only the GSAE uncovers the linear trajectory of the graph indicating that simple embedding of edit distances, WL kernels and other graph autoencoders do not uncover the trajectory as well. Further, we quantify the structure in these embeddings in Figure \ref{fig:toy}B by computing the {\em graph dirichlet energy}  of the signal formed by the sequence index, i.e., the signal $f=[0, 1, \ldots, 10000]$. This is done using the equation for smoothness presented in Section \ref{sec:methods}-C.  Lower values indicate a higher level of smoothness present in the embedding. We see in Figure-\ref{fig:toy}B that aside from a direct embedding of the graph edit distance, GSAE has the highest degree of smoothness.

\begin{figure}[!ht]
    \centering
    \includegraphics[width=0.95\linewidth]{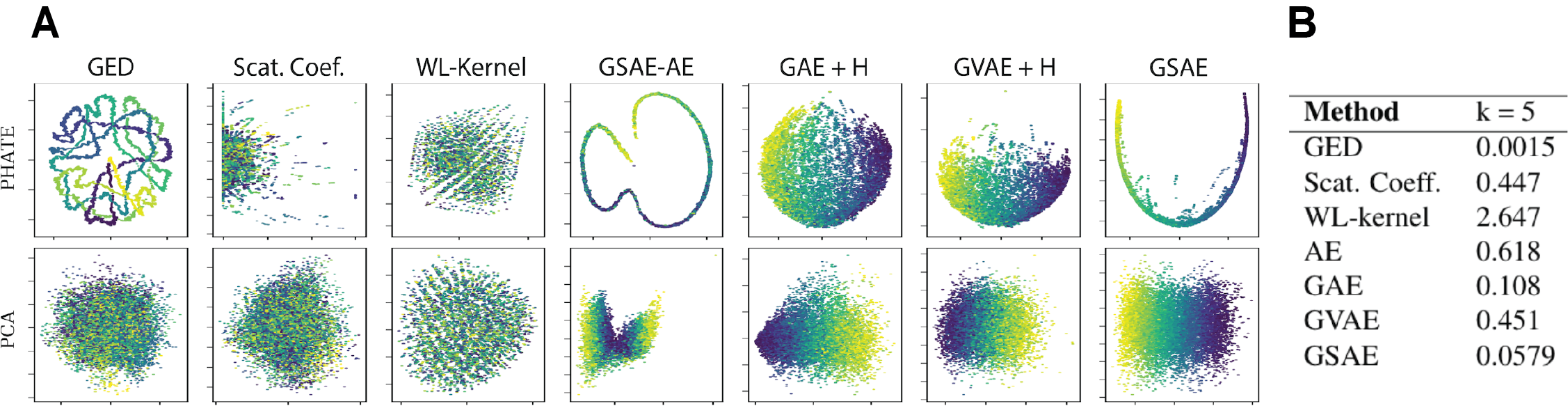}
    \caption[textfont=small]{A. PHATE and PCA plots of seven different embeddings of the random graph dataset.  Color corresponds to the position in the 10,000-step sequence of graphs, the ordering of which GSAE reveals clearly. B. Graph dirichlet energy with respect to the step indices of the trajectory sequence.}
    \label{fig:toy}
\end{figure}

\subsection{RNA Experiments}

We assess the ability of different models to recover RNA structure landscapes and visualize a smooth energy landscape. We train all neural networks with the same penalties, reconstruction as well as energy regression with hyperparameter $\alpha=0.5$ unless otherwise noted.

\subsection{Faithfulness}

Figure~\ref{fig:seq3} contains PHATE and PCA visualizations of SEQ3 embeddings, and shows that only the GSAE model organizes the embeddings by both energy and structure despite using the equally weighted reconstruction and regression penalties. Only the GSAE which recapitulates the bistability of SEQ3 and SEQ4 \cite{hobartner2003bistable} clearly.  

In Figure \ref{fig:density} we show the density plots for the PHATE and PCA plots of GSAE embeddings for each of the four RNA datasets. As described in the paper, we recapitulate the bistable nature of SEQ3 and SEQ4 and visualize this further in the  Figure \ref{fig:density} (top row). In the HIVTAR dataset, we view two clusters of structures rather than the expected single cluster. We hypothesize that this separation may be a result of a minor structural distinction due to the low variability between structures in the HIVTAR dataset. Lastly, we also show that the TEBOWN dataset displays $>$ 2 minima in its density plots (bottom right), which is expected in \cite{cordero2015rich}. Notably, as the energy increases and grows further away from that of the minimum free energy structure, the number of structures possible increases. As a result, instable and structurally diverse folds make up a large portion of RNA folding ensembles.

\begin{figure}[!ht]
    \centering
    \begin{tabular}{cc}
    \includegraphics[width=0.40\linewidth]{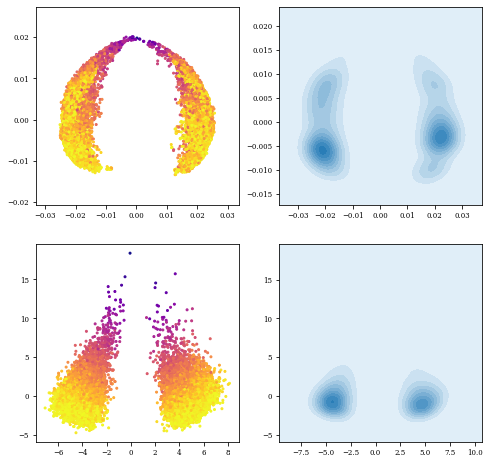} 
    \includegraphics[width=0.40\linewidth]{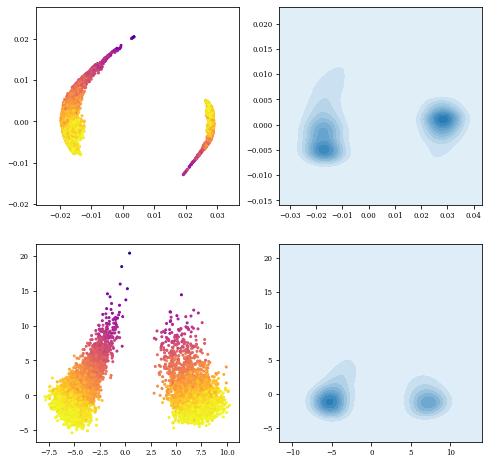}  \\
    \includegraphics[width=0.40\linewidth]{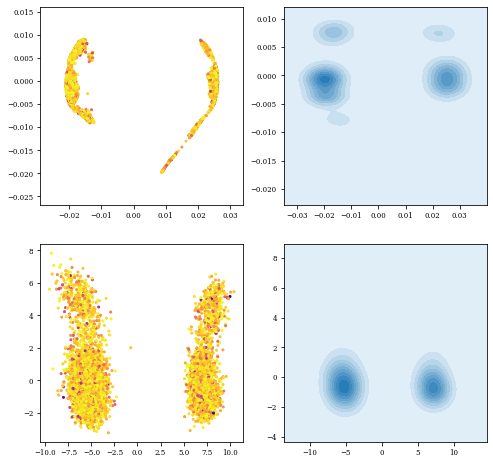}  
    \includegraphics[width=0.40\linewidth]{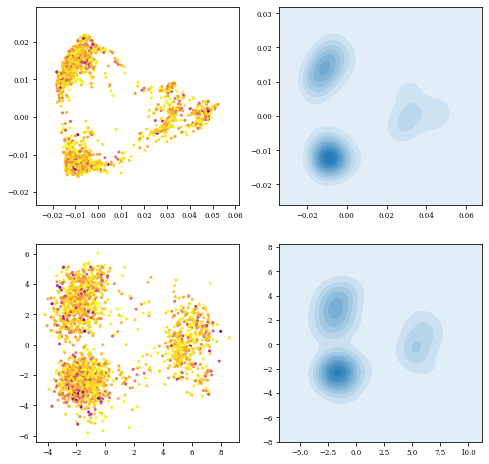}  
    \end{tabular}
    \caption[textfont=small]{Density plots from PHATE and PCA coordinates of RNA embeddings. 25-dimensional embeddings are generated using GSAE and are plotted using PHATE and PCA . The density plot is shown to the right of it's corresponding PHATE and PCA plot. \textit{Top row}: SEQ3, SEQ4. \textit{Bottom Row}: HIVTAR, TEBOWN.}
    \label{fig:density}
\end{figure}

\begin{table}[!ht]
\caption{Results show structural organization of the various embeddings on the two bistable datasets. Graph dirichlet energy with respect to the graph edit distance from the two stable energy minima are reported. Here "+ H" refers to the addition of the energy prediction auxilliary network \textit{H}}
\tiny
\centering
\label{tab:structuralsmooth}
\resizebox{\columnwidth}{!}{%
\begin{tabular}{lllll}
\hline
 & SEQ3 &  & SEQ4 &  \\ \hline
 & Min 1 & Min 2 & Min1 & Min 2 \\
GED & 0.442 $\pm$ 0.0003 & 0.517 $\pm$ 0.0002 & 0.045 $\pm$ 0.0003 & 0.058 $\pm$ 0.0003 \\
Scat. Coeff. & 0.0604 $\pm$ 0.0003 & 0.0732 $\pm$ 0.0002 & 0.066 $\pm$ 0.0002 & 0.0859 $\pm$ 0.0005 \\
GAE & 0.035 $\pm$ 0.001 & 0.045 $\pm$ 0.002 & 0.038 $\pm$ 0.001 & 0.053 $\pm$ 0.003 \\
GAE + H & 0.044 $\pm$ 0.006 & 0.06 $\pm$ 0.006 & 0.043 $\pm$ 0.003 & 0.062 $\pm$ 0.003 \\
GVAE & 0.425 $\pm$ 0.006 & 0.478 $\pm$ 0.008 & 0.443 $\pm$ 0.007 & 0.528 $\pm$ 0.008 \\
GVAE + H & 0.392 $\pm$ 0.005 & 0.46 $\pm$ 0.008 & 0.405 $\pm$ 0.006 & 0.469 $\pm$ 0.006 \\
WL-Kernel & 0.185 $\pm$ 0.0012 & 0.225 $\pm$ 0.001 & 0.2 $\pm$ 0.0016 & 0.263 $\pm$ 0.0016 \\ \hline
\textbf{GSAE - AE} & 0.069 $\pm$ 0.001 & 0.087 $\pm$ 0.002 & 0.069 $\pm$ 0.001 & 0.085 $\pm$ 0.002 \\
\textbf{GSAE (no H)} & 0.337 $\pm$ 0.021 & 0.381 $\pm$ 0.027 & 0.112 $\pm$ 0.004 & 0.038 $\pm$ 0.001 \\
\textbf{GSAE} & 0.346 $\pm$ 0.076 & 0.402 $\pm$ 0.074 & 0.103 $\pm$ 0.004 & 0.124 $\pm$ 0.005 \\ \hline
\end{tabular}%
}
\end{table}

\begin{figure}[!b]
    \centering
    \includegraphics[width=0.95\columnwidth]{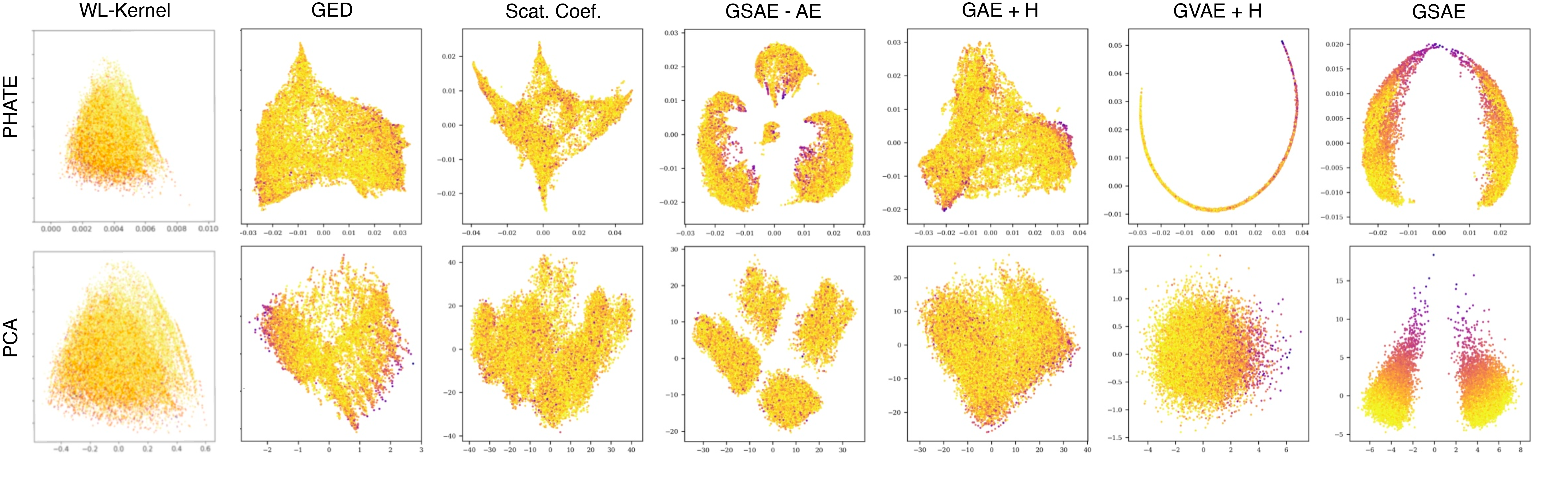}
    \caption[textfont=small]{SEQ3 embedding comparison of various embeddings. SEQ3 is known to be bistable \cite{hobartner2003bistable}, with two energy minima which only GSAE reveals.}
    \label{fig:seq3}
\end{figure}


\subsection{Energy Smoothness}

Energy smoothness quantified for all four RNA sequences in Table~\ref{tab:lap-smooth-energy} and structural smoothness is shown in Table~\ref{tab:structuralsmooth}. While we do not have the ground truth for organizing structures, we show smoothness by graph edit distances to both the bistable minima in SEQ3 and SEQ4, with the idea that as structures move away from these minima, they will also increase in energy. Figure~\ref{fig:density} shows that GSAE can also shed light on the stability landscape of the four RNA structures. SEQ3 and SEQ4 are bistable, while TEBOWN appears to be tristable. However, our embedding shows that HIVTAR can exist in two different fold structures based on the two structures in the embedding, contrary to what is reported in \cite{ganser2019roles}.  

\begin{table}[!ht]
\caption{Graph dirichlet energy of molecule free energy signal over K-NN graph of embedding.  Here "+ H" refers to the addition of the energy prediction network H}
\centering
\label{tab:lap-smooth-energy}
\resizebox{\columnwidth}{!}{%
\begin{tabular}{lllll}
\hline
 & SEQ3 & SEQ4 & HIVTAR & TEBOWN \\ \hline
GED & 0.409 $\pm$ 0.014 & 0.417 $\pm$ 0.031 & 0.105 $\pm$ 0.002 & 0.729 $\pm$ 0.039 \\
Scat. Coeff. & 0.345 $\pm$ 0.009 & 0.390 $\pm$ 0.007 & 0.105 $\pm$ 0.002 & 0.649 $\pm$ 0.025 \\
GAE & 0.331 + 0.008 & 0.345 + 0.008 & \textbf{0.101$\pm$0.002  } & 0.556 + 0.014 \\
GAE + H & 0.128 + 0.006 & 0.096 + 0.007 & 0.102 $\pm$ 0.005 & 0.367 $\pm$ 0.010 \\
GVAE & 0.485 $\pm$ 0.014 & 0.799 $\pm$ 0.018 & 0.124 $\pm$ 0.003 & 0.547$\pm$0.016 \\
GVAE + H & 0.345 $\pm$ 0.009 & 0.276 $\pm$ 0.007 & 0.119 $\pm$ 0.003 & 0.546 $\pm$ 0.014 \\
WL-kernel & 0.636 + 0.048 & 1.091 + 0.083 & 0.185 + 0.013 & 0.559 + 0.033 \\ \hline
\textbf{GSAE - AE } & 0.209 $\pm$ 0.003 & 0.170 $\pm$ 0.002 & \textbf{0.101 $\pm$ 0.001} & 0.435 $\pm$ 0.008 \\
\textbf{GSAE (no H)} & 0.396 $\pm$ 0.011 & 0.444 + 0.007 & 0.105 $\pm$ 0.002 & 0.506 $\pm$ 0.014 \\
\textbf{GSAE } & \textbf{0.105 $\pm$ 0.006} & \textbf{0.081 $\pm$ 0.003} & 0.109$\pm$0.002 & \textbf{ 0.352 $\pm$ 0.026} \\ \hline
\end{tabular}
}
\end{table}

\subsection{Energy Prediction}

We show the energy prediction accuracy of the models at various settings of the parameter $\alpha$ in Table \ref{tab:energy_prediction} which decides the penalty balance between the autoencoding reconstruction penalty and the energy prediction penalty. Note that here, that despite training all other models with the auxiliary energy prediction task (as indicated by the network+H label), we see that the some of the models such as GVAE have decreased accuracy, potentially because of loss of information in the feature encodings of the embedding layers. We note that GSAE is able to simultaneously organize the embedding by structure and energy such that energy can be successfully predicted. We also note that the GAE is able to predict energy as well. However, as indicated by Table \ref{tab:lap-smooth-energy}, it does not maintain structural information relating the metaproperty of interest as well as the GSAE in it's embedding.

\begin{table}[!ht]
\caption{Performance of auxiliary network H. Energy prediction MSE (mean $\pm$ std.\ over 10 runs) on each of the four RNA datasets}
\centering
\resizebox{\columnwidth}{!}{%
\begin{tabular}{lllll}
\toprule
& SEQ3 & SEQ 4 & HIVTAR & TEBOWN \\ \midrule
GAE & 224.832 $\pm$ 291.277 & 360.797 $\pm$ 416.404 & 217.451 $\pm$ 190.157 & 168.191 $\pm$ 205.224 \\
GAE + H ($\alpha=0.1$) & 1.223 $\pm$ 0.069 & 1.364 $\pm$ 0.119 & 3.159 $\pm$ 0.090 & 0.624 $\pm$ 0.031 \\
GAE + H ($\alpha=0.5$) & 1.247 $\pm$ 0.0084 & 1.377 $\pm$ 0.101 & 3.174 $\pm$ 0.078 & 0.608 $\pm$ 0.025 \\
GVAE & 99.442 $\pm$ 7.386 & 156.922 $\pm$ 10.508 & 207.148 $\pm$ 12.742 & 10.028 $\pm$ 2.431 \\
GVAE + H ($\alpha=0.1$) & 5.536 $\pm$ 0.089 & 6.996 $\pm$ 0.234 & 3.168 $\pm$ 0.045 & 0.741 $\pm$ 0.021 \\
GVAE + H ($\alpha=0.5$) & 4.338 $\pm$ 0.0789 & 5.625 $\pm$ 0.434 & 3.188 $\pm$ 0.037 & 0.750 $\pm$ 0.015 \\ \midrule
\textbf{GSAE - AE} & 2.875 $\pm$ 0.04 & 3.877 $\pm$ 0.053 & 3.176 $\pm$ 0.044 & 0.678 $\pm$ 0.01 \\
\textbf{GSAE (no H)} & 98.561 $\pm$ 3.35 & 156.567 $\pm$ 4.292 & 209.654 $\pm$ 8.425 & 8.930 $\pm$ 2.948 \\
\textbf{GSAE ($\alpha=0.1$)} & 1.786 $\pm$ 0.639 & 2.908 $\pm$ 0.788 & 3.739 $\pm$ 0.477 & 0.722 $\pm$ 0.008 \\
\textbf{GSAE ($\alpha=0.5$)} & 1.795 $\pm$ 0.533 & 2.040 $\pm$ 0.587 & 3.509 $\pm$ 0.201 & 0.661 $\pm$ 0.246\\
\bottomrule
\end{tabular}%
}
\label{tab:energy_prediction}
\end{table}

\subsection{Generating Folding Trajectories with SIN}

\begin{figure}[!ht]
    \centering
    \begin{tabular}{cc}\includegraphics[width=0.17\linewidth]{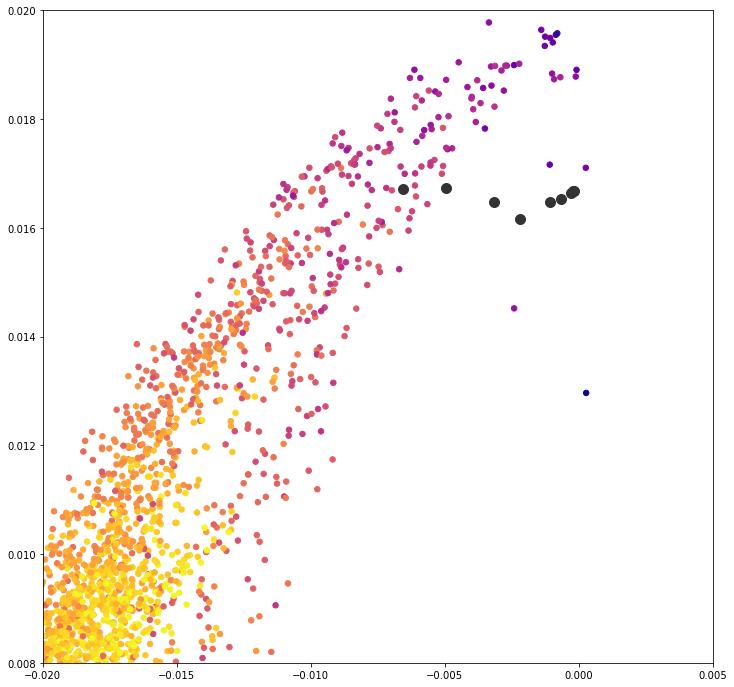} &
    \includegraphics[width=0.70\linewidth]{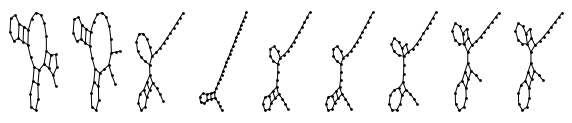}
    \end{tabular}
    \caption[textfont=small]{Example trajectory from the PHATE embedding of the GSAE latent space and the corresponding RNA graphs.}
    \label{fig:trajectory}
\end{figure}

We also emphasize that the GSAE is a generative model, trained as a VAE, therefore we can sample trajectories of folds in the landscape as potential paths from high to low energy folds. This is depicted on a sample trajectory in Figure~\ref{fig:trajectory}. We present quantitative  reconstruction results of the inverse model described in Section \ref{scat-inverse} in Table \ref{table:inv-quant}.
This model can be used in a generative setting to produce sequences of graphs that resemble RNA folding trajectories.  To achieve this we first train a GSAE model with small latent space dimension over RNA graphs from one of the datasets.  Then for two randomly chosen RNA graphs in the dataset we sample from the line segment connecting their corresponding latent space embeddings.  These interpolated points in the latent space are mapped into the space of scattering coefficients by the decoding network of the GSAE.  Finally these points in scattering coefficient space are fed into the inverse model SIN.  The weights of the resulting adjacency matrices are rounded to produce unweighted graphs.

To see this method in action we trained the GSAE model with latent space dimension 5 on 70,000 graphs from the SEQ3 dataset.  In selecting the end points for our generative trajectories, we sampled the starting graph from the subset of high-energy configurations and the final graph from the low-energy configurations.  See Figure \ref{fig:gen-trajectories}A for trajectories generated using this method.  In Figure \ref{fig:gen-trajectories}B for every trajectory we compute the graph edit distance between the final graph and each individual graph in the trajectory.  The results suggest that in most cases, these generative trajectories are smooth in terms of graph edit distance.

\begin{table}[!ht]
\centering
\caption{Inverse model test set reconstruction error generating adjacency matrices from scattering coefficients over N=10 runs}
\label{table:inv-quant}
\begin{tabular}{ll}

 \begin{tabular}{ll}
 \toprule
 {} &      MSE $\pm$ std $\times 10^{-3}$\\
 \midrule
 SEQ3   &  0.070 $\pm$ 0.010 \\
 SEQ4   &  0.059 $\pm$ 0.004 \\
 HIVTAR &  7.425 $\pm$ 2.459 \\
 TEBOWN &  7.175 $\pm$ 3.552 \\
 \bottomrule
 \end{tabular}
 \end{tabular}
 \end{table}

\begin{figure}[!ht]
    \centering
    \begin{tabular}{cc}
    \includegraphics[width=0.95\columnwidth]{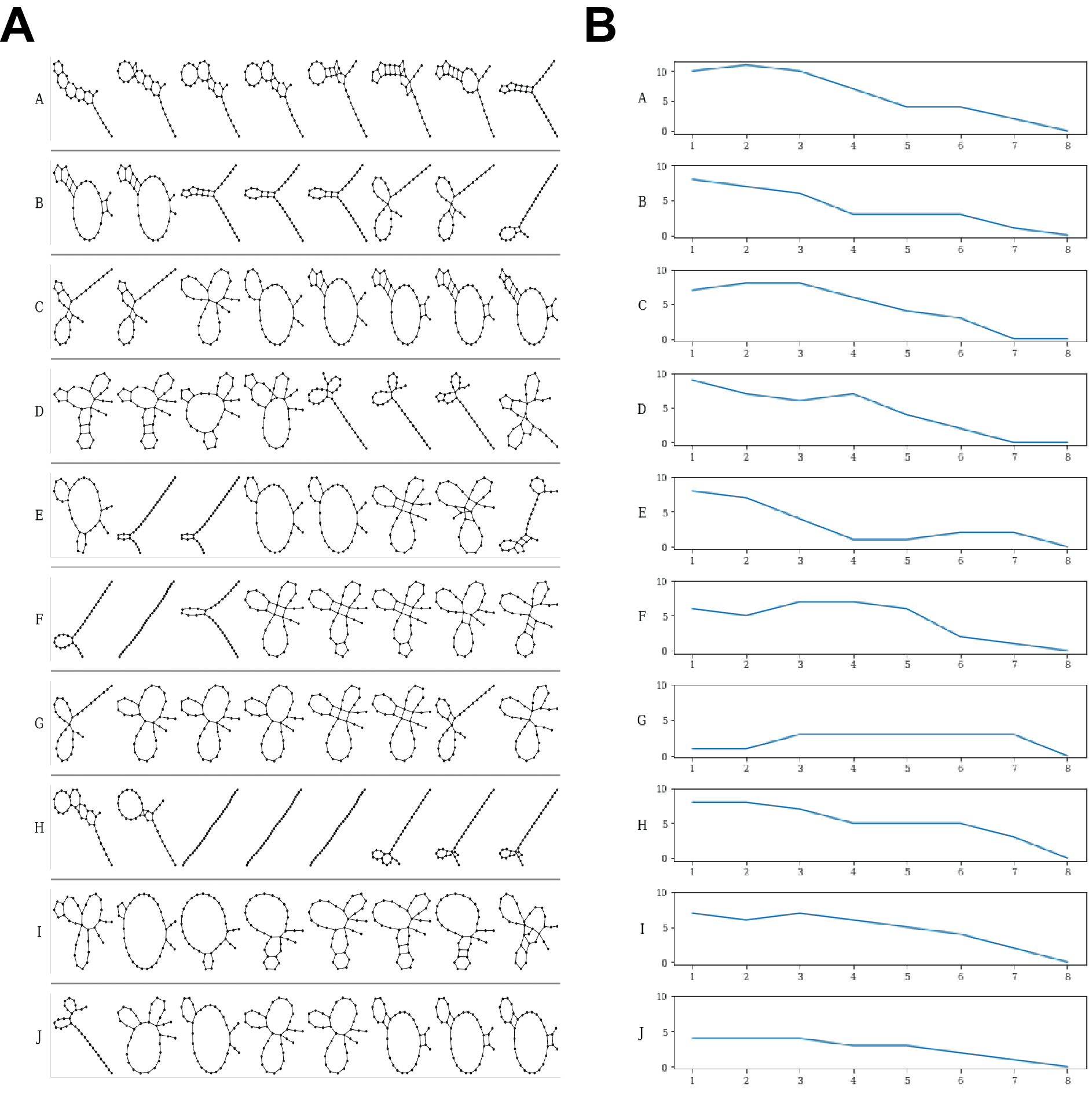} 
    \end{tabular}
    \caption[textfont=small]{A. Sample trajectories produced by applying the scattering inverse network to linear interpolations between training points in GSAE latent space. B. Edit distance between each individual graph in the trajectory and the final graph in the trajectory for all the trajectories}
    \label{fig:gen-trajectories}
\end{figure}



 




\subsection{Pooling Method Comparison for Graph Embeddings}

\begin{figure}[!ht]
    \centering
    \begin{tabular}{cc}
    \includegraphics[width=0.9\columnwidth, height=4cm]{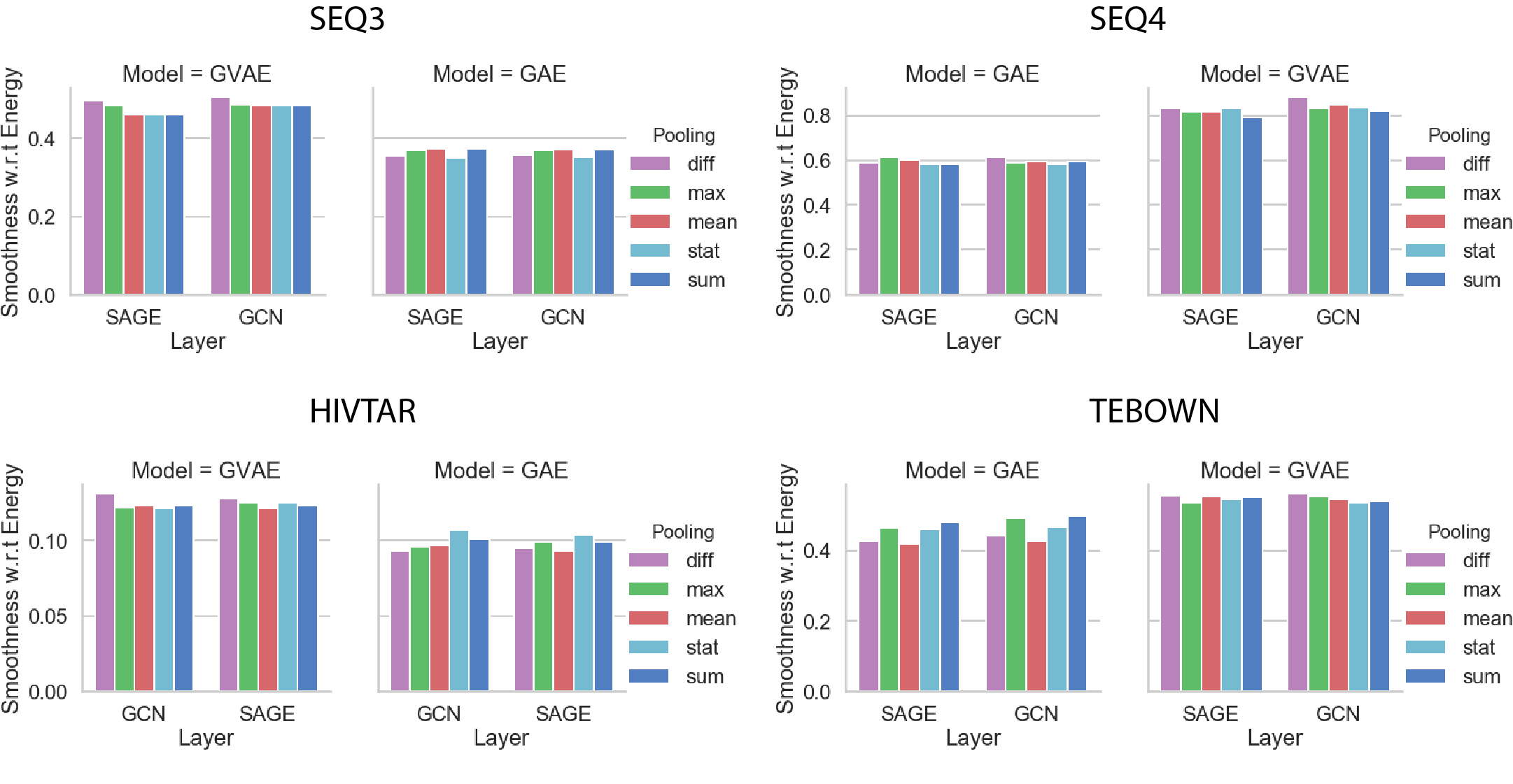} 
    \end{tabular}
    \caption[textfont=small]{We compare the effect of different pooling methods  on ability of the GAE and GVAE models to enforce smoothness in the embedding with respect to energy, quantified by the smoothness index described in Section 3.}
    \label{fig:pooling-smooth}
\end{figure}

\begin{figure}[!ht]
    \centering
    \begin{tabular}{cc}
    \includegraphics[width=0.9\columnwidth]{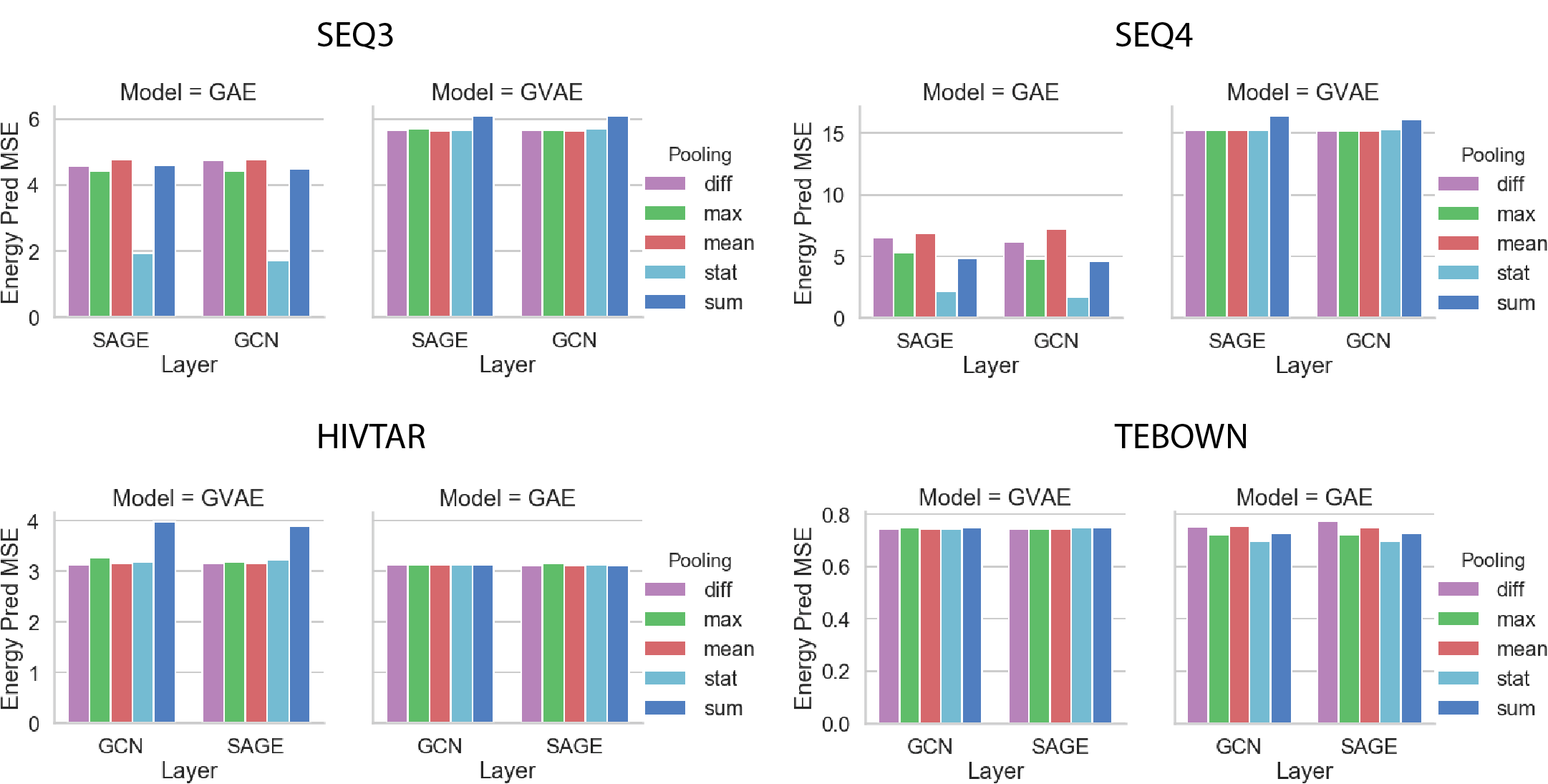} 
    \end{tabular}
    \caption[textfont=small]{Here the effect of different pooling methods on the GAE and GVAE model's ability to predict energy, measured by test MSE.}
    \label{fig:pooling-mse}
\end{figure}

As the geometric scattering construction produces graph-level representation through statistical moments, we examine how these pooling approach performs in the GAE and GVAE models relative to more traditional pooling methods. We also try two different GNN layer types, GCN and GraphSAGE for this experiment. From Figure \ref{fig:pooling-smooth} and Figure \ref{fig:pooling-mse}, observe that that statistical moment pooling achieves similar performance to other pooling methods. For the SEQ3 and SEQ4 datasets, we observe that statistical moment pooling achieves better performance on both the energy smoothness metric and the energy prediction task. We rationalize this effect as a result of the pooling method's ability to capture additional information about a signal's distribution across nodes.

\section{Conclusion} 

We have presented a hybrid approach that combines geometric scattering transforms on graphs with variational autoencoders in order to embed biomolecular graphs. We propose this as a useful way of learning a large multi-scale set of features that are descriptive of graph structure which are then reweighted and combined appropriately using a trained autoencoder. In this way we can complement the representational power of graph scattering transforms with the flexibility of trainable autoencoder embeddings to achieve the properties desired in a low-dimensional space. Furthermore we go beyond only embedding graphs present in our training dataset by training our architecture as a generative model. A generative-based approach is especially powerful in the data domain of biomolecules due to the enormous space of structures possible. For example, the smallest of the sequences used in our study, SEQ3, can theoretically fold into $10^8$ possible structures. For most RNA sequences of interest, exhaustive enumeration of possible folds is intractable so the ability to sample nearby folds not present in initial dataset allows for exploration of structural space not before possible. Morever, in this work we demonstrate the utility of our proposed approach to recover folding landscapes of functional RNA however this framework can also be applied to other biomolecular data domains. The structure-centric nature of biomolecules presents an opportunity to leverage graph-based approaches such as ours to better study datasets from these domains.


\bibliographystyle{IEEEtran}
\bibliography{main.bib}

\end{document}